%% file: 0main.tex
\documentclass[sigconf]{acmart}
\usepackage{enumitem}
\usepackage{multirow} 
\usepackage{booktabs} 
\usepackage{tabularx} 
\usepackage{ragged2e} 
\usepackage{graphicx}
\usepackage{subcaption} 
\usepackage{algorithm}
\usepackage{algorithmic}
\usepackage{amsthm}
\usepackage{enumitem}
\usepackage[table]{xcolor} 
\usepackage{amsmath}       
\usepackage{balance}

\AtBeginDocument{%
  }

\copyrightyear{2026}
\acmYear{2026}
\setcopyright{cc}
\setcctype{by}
\acmConference[CIKM '26]{Proceedings of the 35th ACM International Conference on Information and Knowledge Management}{November 07--11, 2026}{Rome, Italy}
\acmBooktitle{Proceedings of the 35th ACM International Conference on Information and Knowledge Management (CIKM '26), November 07--11, 2026, Rome, Italy}
\acmDOI{10.1145/3799682.3841113}
\acmISBN{979-8-4007-2539-5/2026/11}

\begin{document}

\title{PRO-Bid: Pareto-Prioritized Regret Optimization for Constraint-Aware Generative Auto-Bidding}

\author{Binglin Wu}
\email{binglin@mail.dlut.edu.cn}
\orcid{0009-0000-2295-6773}
\authornotemark[1]
\affiliation{%
  \institution{Dalian University of Technology}
  \city{Dalian}
  \country{China}
}
\affiliation{%
  \institution{National Key Laboratory of Maritime Decision Intelligence}
  \city{Dalian}
  \country{China}
}

\author{Yingyi Zhang}
\email{yingyizhang@mail.dlut.edu.cn}
\orcid{0000-0001-9062-3428}
\affiliation{%
  \institution{Dalian University of Technology}
  \city{Dalian}
  \country{China}
}
\affiliation{%
  \institution{City University of Hong Kong}
  \city{Hong Kong}
  \country{China}
}

\author{Xianneng Li}
\email{xianneng@dlut.edu.cn}
\orcid{0000-0003-4130-6930}
\correspondingauthor
\affiliation{%
  \institution{Dalian University of Technology}
  \city{Dalian}
  \country{China}
}
\affiliation{%
  \institution{National Key Laboratory of Maritime Decision Intelligence}
  \city{Dalian}
  \country{China}
}

\author{Ruyue Deng}
\email{dengruyue.dry@alibaba-inc.com}
\affiliation{%
  \institution{Alibaba International Digital
Commerce Group}
  \city{Hangzhou}
  \country{China}
}

\author{Chuan Yue}
\email{yuechuan.yc@alibaba-inc.com}
\correspondingauthor
\affiliation{%
  \institution{Alibaba International Digital
Commerce Group}
  \city{Hangzhou}
  \country{China}
}

\author{Weiru Zhang}
\email{weiru.zwr@alibaba-inc.com}
\affiliation{%
  \institution{Alibaba International Digital
Commerce Group}
  \city{Hangzhou}
  \country{China}
}

\author{Xiaoyi Zeng}
\email{yuanhan@taobao.com}
\affiliation{%
  \institution{Alibaba International Digital
Commerce Group}
  \city{Hangzhou}
  \country{China}
}

\renewcommand{\shortauthors}{Binglin Wu et al.}


\begin{abstract}
Auto-bidding systems strive to maximize marketing value while maintaining high compliance with efficiency constraints, such as Target Cost-Per-Action (CPA). While Decision Transformers offer powerful sequence modeling capabilities, their application to this setting faces two challenges: 1) standard Return-to-Go conditioning causes state aliasing by ignoring the cost dimension, preventing precise resource pacing; and 2) standard regression constrains the policy to mimic historical averages, limiting its capacity to optimize performance near the high-efficiency boundary. To tackle these challenges, we propose PRO-Bid, a constraint-aware generative auto-bidding framework featuring systematic redesigns across data, architecture, and training via two synergistic mechanisms: 1) Constraint-Decoupled Pareto Representation (CDPR) separates global constraints into recursive cost and value contexts to restore resource perception, while reweighting trajectories based on the empirical Pareto frontier to prioritize high-efficiency data; and 2) Counterfactual Regret Optimization (CRO) employs a global predictor to evaluate alternative actions and identify promising local adjustments. By utilizing these high-utility outcomes as weighted regression targets, the model overcomes mean regression and approaches the empirical high-efficiency boundary. Extensive experiments on two public benchmarks and online A/B tests show that PRO-Bid achieves better constraint satisfaction and value acquisition than state-of-the-art baselines.
\end{abstract}



\maketitle

\footnotetext[1]{Work is done during the internship at Alibaba International Digital Commerce Group.}

\input{chapter/1intro}
\input{chapter/3method}
\input{chapter/4expr}
\input{chapter/5online}
\input{chapter/2related}
\input{chapter/6conclu}

\appendix
\input{chapter/7appendix}

\begin{acks}
This work was supported by the National Natural Science Foundation of China (NSFC) under Grant 72071029 and 72231010, and the Graduate Research Fund of the School of Economics and Management of Dalian University of Technology (No. DUTSEMDRFKO1).
\end{acks}

\section*{GenAI Usage Disclosure}
During the preparation of this manuscript, the authors utilized Gemini 3.0 Pro exclusively for the purposes of improving readability, grammar, and spelling. No generative AI tools were used to formulate the research ideas, design the methodology, conduct the experiments, or generate the core scientific claims. 

\bibliographystyle{ACM-Reference-Format}
\balance
\bibliography{chapter/8ref}

\end{document}

%% file: chapter/1intro.tex
\section{Introduction}

Auto-bidding serves as the backbone of modern computational advertising~\cite{zhao2018deep, balseiro2021robust, ou2023deep, li2025cusmer, zhang2026evoking, zhang2026personalize}, processing billions of Real-Time Bidding (RTB) decisions daily. This mechanism is formulated as a constrained sequential decision-making problem, in which an agent dynamically adjusts bids for impression opportunities over a fixed campaign period~\cite{li2024trajectory, aggarwal2024auto, zhang2025adapting}. The primary objective is to maximize cumulative marketing value~\cite{borissov2010automated}, such as Gross Merchandise Value (GMV) or conversion volume, while maintaining high compliance with efficiency constraints. These constraints are typically defined by period-level ratio metrics, such as Cost-Per-Click (CPC)~\cite{zhu2017optimized}, Cost-Per-Action (CPA)~\cite{nazerzadeh2008dynamic,hu2016incentive}, or Return on Ad Spend (ROAS)~\cite{deng2023multi}. Consequently, balancing value maximization with high adherence to cost constraints in dynamic markets remains a central challenge for auto-bidding systems.

To mitigate the substantial financial risks associated with online exploration, offline Reinforcement Learning (RL)~\cite{korenkevych2024offline, li2024trajectory, he2024hierarchical} and sequence modeling~\cite{jiang2025optimal,peng2025expert} have emerged as prevalent approaches. These methods aim to learn policies directly from historical bidding logs. However, advertising delivery constitutes a complex long-horizon resource allocation problem~\cite{guo2024generative,duan2025adaptable}. Optimal strategies typically reside on the boundary of the feasible region~\cite{asadpour2019concise, ou2023survey}, characterized by full budget utilization and high adherence to target ratio constraints. Consequently, effective models require global planning capabilities rather than solely maximizing the immediate rewards of individual bids. Specifically, agents must dynamically adjust bidding strategies based on real-time resource pacing to precisely achieve campaign performance goals.

Recently, the Decision Transformer (DT)~\cite{chen2021decision,li2025gas,gao2025generative} has been employed in auto-bidding tasks to capture long-range dependencies. Nevertheless, existing methods encounter two fundamental limitations when addressing hard ratio constraints. \textbf{1) Standard Return-to-Go (RTG) conditioning induces state aliasing under ratio constraints.} Although RTG encodes future cumulative value, it neglects the cumulative cost required to realize it. Ratio-based metrics, such as CPA, inherently depend on both cost and value dimensions. Relying solely on remaining value provides insufficient information, precluding the model from distinguishing states with varying cost efficiencies. This deficiency in perceiving resource dynamics impedes precise pacing. \textbf{2) Standard regression objectives restrict the policy to historical averages.} Prevailing generative models minimize the Mean Squared Error (MSE) uniformly across all historical actions. This objective compels the policy to approximate the average data distribution, mimicking both high-performing strategies and suboptimal behaviors, including constraint violations. Absent a mechanism to prioritize superior outcomes, the model struggles to consistently approach the high-efficiency regions of the data support. Consequently, the agent fails to effectively approach the empirical limit of the feasible region.

To address these challenges, we propose \textbf{PRO-Bid} (Auto-\textbf{bid}ding via \textbf{P}areto-Prioritized \textbf{R}egret \textbf{O}ptimization). This DT-based framework incorporates two synergistic mechanisms: \textbf{1) Constraint-Decoupled Pareto Representation (CDPR)}, which enhances standard RTG conditioning by decomposing global ratio constraints into two recursive state streams: \textit{Remaining Required Value} and \textit{Remaining Allowable Cost}. This decoupled structure enables real-time perception of marginal resource changes. Furthermore, historical trajectories are reweighted based on their proximity to the empirical Pareto frontier to focus learning on high-efficiency data, enhancing representation quality at the input level. \textbf{2) Counterfactual Regret Optimization (CRO)}, which formulates active policy improvement as a robust supervised learning task. Rather than passively mimicking historical data, this module employs a predictor to evaluate alternative actions within the policy's local variance bound, identifying promising adjustments to the current strategy. These refined targets then serve as high-quality pseudo-labels. By regressing the policy toward these weighted targets, PRO-Bid progressively shifts the bidding strategy toward the Pareto frontier, aiming to maximize value acquisition while managing constraint violation risks. Our contributions are summarized as follows:
\begin{itemize}[leftmargin=*]
    \item We propose PRO-Bid, a constraint-aware generative auto-bidding framework. To effectively handle hard ratio constraints, it introduces systematic improvements across three dimensions of sequence modeling: data, architecture, and training.
    \item Data \& Architecture enhancements via CDPR: CDPR prioritizes high-quality trajectories by using an empirical Pareto frontier to reweight suboptimal logs. Additionally, it resolves state aliasing by decoupling the global constraint into a dual-stream context to restore precise resource perception.
    \item Training enhancement via CRO: Moving beyond standard behavior cloning, CRO utilizes a global predictor to identify high-utility local adjustments. By treating these as weighted regression targets, it actively drives the policy toward the empirical high-efficiency frontier.
    \item Extensive experiments on two public benchmarks and online A/B tests in a large-scale advertising system show that PRO-Bid achieves superior constraint satisfaction and value acquisition.
\end{itemize}

%% file: chapter/3method.tex
\section{Preliminaries}
\subsection{Problem Formulation of Auto-Bidding}

We formulate the auto-bidding task as a constrained optimization problem widely adopted in industry advertising systems~\cite{guo2024generative}. Consider an advertiser participating in $n$ impression auctions over a fixed period. For each impression $i$ with private valuation $v_i$ and potential cost $c_i$, the objective is to maximize the total accumulated value while adhering to both a budget constraint and a global efficiency ratio constraint:
\begin{equation}
\label{eq:optimization_problem}
\begin{aligned}
\text{maximize} & \quad \sum_{i=1}^{n} x_i v_i \\
\text{s.t.} & \quad \sum_{i=1}^{n} c_ix_i \le B \\
& \quad \frac{\sum_{i=1}^{n} c_i x_i}{\sum_{i=1}^{n} p_i x_i} \le \rho_{\text{tgt}} \\
& \quad x_i  \in \{0, 1\}, \quad \forall i 
\end{aligned}
\end{equation}
where $x_i$ indicates whether impression $i$ is won. Under this formulation, the system enforces the total budget $B$ alongside a specific ratio-based constraint target $\rho_{\text{tgt}}$. The term $p_i$ represents the constraint-specific indicator. In a target CPA campaign, for instance, $p_i$ denotes the conversion indicator, $v_i$ is the conversion value, and $\rho_{\text{tgt}}$ represents the target CPA ratio.

Directly solving this combinatorial problem is intractable. According to primal-dual theory~\cite{he2021unified}, the problem can be reformulated by introducing Lagrange multipliers or shadow prices. The optimal bid $b_i^*$ is then derived as a function of these global parameters:
\begin{equation}
b_i^* = \lambda_0 v_i + \lambda_1 p_i \rho_{\text{tgt}}
\end{equation}

In the common advertising scenario where the optimization objective aligns with the constraint metric, such as maximizing conversions under a CPA constraint where $p_i = v_i$, the bidding formula mathematically simplifies to a proportional bidding strategy:
\begin{equation}
\label{eq:proportional_bidding}
b_i^* = \lambda^* v_i
\end{equation}
In this context, $\lambda^*$ serves as a unified pacing parameter that governs the trade-off between value acquisition and constraint adherence. Consequently, the auto-bidding task is recast as learning a sequential policy $\pi$ that outputs the appropriate $\lambda_t$ at each step $t$ to satisfy the global ratio constraint.

\subsection{Generative Bidding via DT}

Within the DT framework~\cite{chen2021decision}, the auto-bidding process is modeled as a conditional sequence generation task. A trajectory $\tau$ is represented as a sequence of tokens:
\begin{equation}
\label{eq:trajectory}
\tau = \left( R_1, s_1, a_1, \dots, R_T, s_T, a_T \right)
\end{equation}
The key components are defined as follows:
\begin{itemize}[leftmargin=*]
    \item \textbf{State ($s_t$):} A context vector encapsulating campaign status, including elapsed time and historical features.
    \item \textbf{Action ($a_t$):} The continuous bidding parameter $\lambda_t \in \mathbb{R}^+$ applied at step $t$.
    \item \textbf{Return-to-Go ($R_t$):} The target cumulative reward from step $t$, defined recursively as $R_t = \sum_{t'=t}^{T} r_{t'}$.
\end{itemize}

The model learns a deterministic policy $\pi_\theta$ to predict action $a_t$ by minimizing the MSE over the offline dataset $\mathcal{D}$:
\begin{equation}
\label{eq:mse}
\mathcal{L}_{DT} = \frac{1}{T} \sum_{t=1}^{T} \left( a_t - \pi_\theta(R_{\leq t}, s_{\leq t}, a_{<t}) \right)^2
\end{equation}

During inference, the generation process is conditioned on a desired total return $R_{1}$, and the token $R_t$ is updated deterministically based on realized rewards.

However, this framework faces two critical issues when applied to the ratio-constrained problem defined in Eq.~\eqref{eq:optimization_problem}.
\begin{itemize}[leftmargin=*]
    \item \textbf{The trajectory formulation in Eq.~\eqref{eq:trajectory} induces state aliasing.} Since global ratio constraints depend on both accumulated cost and accumulated value, conditioning solely on remaining value $R_t$ fails to encode the remaining allowable cost, which prevents the agent from perceiving precise resource boundaries.
    \item \textbf{The MSE objective in Eq.~\eqref{eq:mse} enforces mean regression.} By fitting the average behavior of suboptimal historical logs, the policy lacks explicit gradients to actively approach the empirical high-efficiency boundary.
\end{itemize}

\section{Framework}

In this section, we introduce PRO-Bid, a framework that integrates two synergistic modules for constrained bidding, as illustrated in Figure 1. Rather than making isolated adjustments to standard Decision Transformers, PRO-Bid systematically reconstructs the sequence modeling process across three levels: data, architecture, and training.
First, CDPR (Sec. 3.1) operates at both the data and architecture levels: it prioritizes high-efficiency data through Pareto reweighting and resolves state aliasing via dual-stream recursive contexts. Second, CRO (Sec. 3.2) modifies the training objective. It leverages a global predictor to identify high-utility local adjustments as regression targets, enabling the policy to overcome mean regression and approach the empirical high-efficiency boundary.

\begin{figure*}[th]
    \centering
    \includegraphics[width=0.98\linewidth]{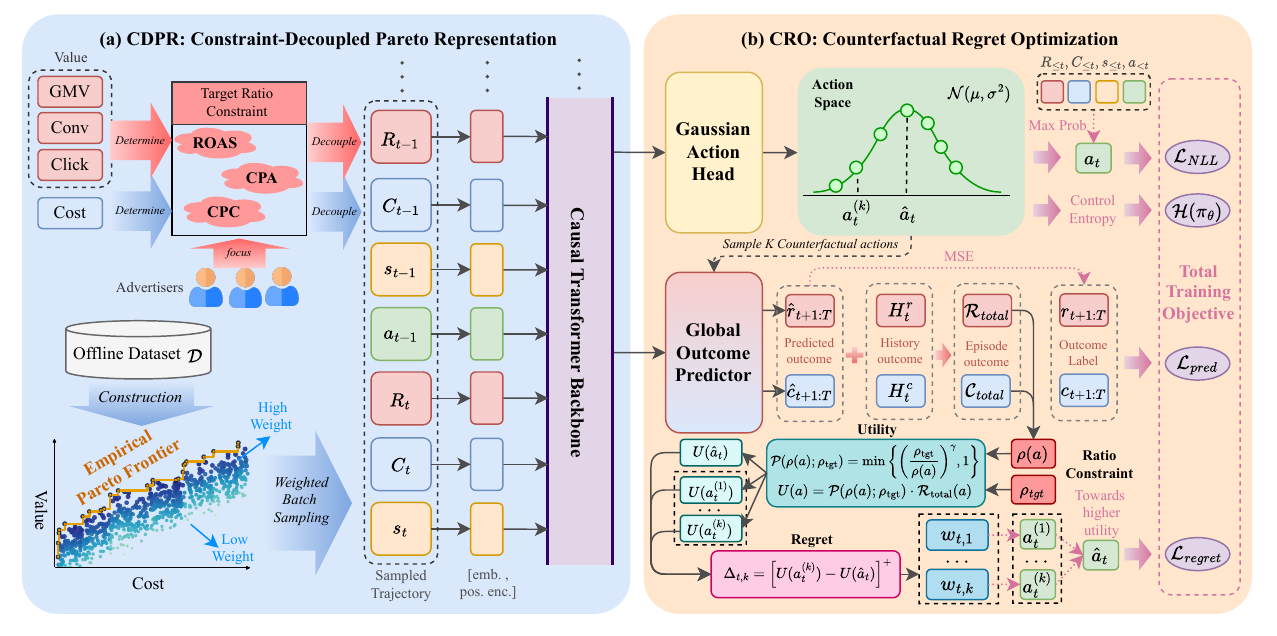}
    \caption{Overall framework of PRO-Bid.}
    \label{fig:framework}
\end{figure*}

\subsection{Constraint-Decoupled Pareto Representation}
\label{sec:module1}

The explicit perception of resource boundaries relative to campaign objectives is a fundamental prerequisite for the satisfaction of hard constraints. To achieve this, we introduce a dual-stream context construction that resolves state aliasing, complemented by a Pareto-prioritized reweighting mechanism that calibrates the data distribution.

\subsubsection{\textbf{Dual-Stream Context Construction}}

To resolve state aliasing where an agent fails to perceive resource consumption, we reconstruct the conditioning context to explicitly track boundary dynamics. We decouple the campaign objective into two additive, recursive sequences: Return-to-Go ($R_t$) and Cost-to-Go ($C_t$).

We define the tokens at time step $t$ as:
\begin{equation}
    R_t = \sum_{t'=t}^{T} r_{t'}, \quad C_t = \sum_{t'=t}^{T} c_{t'}
\end{equation}
where $r_{t'}$ and $c_{t'}$ denote the value and cost at step $t'$, respectively. The recursive transition is defined as:
\begin{equation}
    R_{t+1} = R_t - r_t, \quad C_{t+1} = C_t - c_t
\end{equation}
We augment the input trajectory $\tau$ with this dual-stream context:
\begin{equation}
    \label{eq:aug}
    \tau_{aug} = (R_1, C_1, s_1, a_1, \dots, R_T, C_T, s_T, a_T)
\end{equation}
By conditioning the policy as $\pi_\theta(a_t | R_{\leq t}, C_{\leq t}, s_{\leq t}, a_{<t})$, the model learns the joint distribution of costs and rewards. This decoupled formulation mathematically generalizes to any ratio-based efficiency metric. For instance, while a Target CPA tracks required conversions and allowable cost, a Target ROAS constraint simply redefines the value stream $R_t$ to track required GMV.  This formulation enables the agent to learn an explicit trade-off function that determines the appropriate bid parameter based on the remaining budget $C_t$ required to acquire value $R_t$.

During online inference, this decoupled formulation functions as a dynamic pacing mechanism. We initialize $C_1 = B$ and $R_1 = B / \rho_{\text{tgt}}$, where $\rho_{\text{tgt}}$ represents the global efficiency target. At each step, the tokens are updated based on realized feedback. If the agent overspends, $C_{t+1}$ depletes faster than $R_{t+1}$. Consequently, the policy operates within a tightened resource boundary and naturally shifts probability mass towards conservative bids, ensuring the campaign adheres to global constraints over the long horizon.

\subsubsection{\textbf{Pareto-Prioritized Experience Reweighting}}

Real-world advertising logs are inherently noisy and contain numerous suboptimal trajectories. Uniform sampling from such a skewed distribution hinders the learning of efficient policies. To address this issue, we assign each trajectory $\tau_i$ a sampling probability based on three unified quality metrics.

\textbf{Empirical Pareto Frontier Construction.}
Advertisers frequently adjust ratio-based efficiency constraints in dynamic markets. Consequently, achieving maximum value for a given cost level inherently maximizes the probability of satisfying these varying targets. Motivated by this, we represent each trajectory $\tau_i$ in the offline dataset $\mathcal{D}$ as a point in the objective space $\mathcal{O}_i = (\mathcal{R}(\tau_i), \mathcal{C}(\tau_i))$. After normalizing these coordinates to $(\tilde{r}_i, \tilde{c}_i)$, a trajectory $\tau_i$ is defined as \textit{Pareto-dominated}~\cite{van2014multi} if there exists another trajectory $\tau_j$ such that $\tilde{c}_j \le \tilde{c}_i$ and $\tilde{r}_j \ge \tilde{r}_i$, with at least one strict inequality. We define the Empirical Pareto Frontier $\mathcal{F}$ as the set of non-dominated trajectories within the offline dataset $\mathcal{D}$. This frontier serves as a proxy for high-efficiency behaviors, guiding the model to prioritize informative samples.

\textbf{Multi-Factor Quality Scoring.}
We quantify the quality of each trajectory $\tau_i$ using three scoring components:

1) \textit{Efficiency Score ($S_{\text{eff}}^{(i)}$):} We encourage the model to learn from trajectories close to the empirical Pareto frontier. We calculate the Euclidean distance $d(\tau_i, \mathcal{F})$~\cite{dokmanic2015euclidean} from trajectory $\tau_i$ to its nearest neighbor on the frontier $\mathcal{F}$. The score is derived using an exponential kernel:
\begin{equation}
    S_{\text{eff}}^{(i)} = \exp\left(-\kappa \cdot d(\tau_i, \mathcal{F})\right)
\end{equation}
where $\kappa$ is a hyperparameter controlling decay sharpness.

2) \textit{Compliance Score ($S_{\text{com}}^{(i)}$):} We apply a soft penalty based on constraint violation. Let $\rho_i = \mathcal{C}(\tau_i) / \mathcal{R}(\tau_i)$ denote the realized efficiency ratio of trajectory $i$, and let $\rho_{\text{tgt}}$ denote the global constraint target. The score is defined as:
\begin{equation}
    S_{\text{com}}^{(i)} = 
    \begin{cases} 
    1 & \text{if } \rho_i \le \rho_{\text{tgt}} \\
    \left( \frac{\rho_{\text{tgt}}}{\rho_i} \right)^\omega & \text{if } \rho_i > \rho_{\text{tgt}}
    \end{cases}
\end{equation}
where $\omega \ge 1$ determines the severity of the penalty.

3) \textit{Richness Score ($S_{\text{len}}^{(i)}$):} We normalize the trajectory length $T_i$ against the maximum episode length $T_{\max}$:
\begin{equation}
    S_{\text{len}}^{(i)} = \frac{T_i}{T_{\max}}
\end{equation}

\textbf{Weighted Batch Sampling.}
The final sampling quality score $\mathcal{Q}_i$ is the product of these components:
\begin{equation}
    \mathcal{Q}_i = S_{\text{eff}}^{(i)} \cdot S_{\text{com}}^{(i)} \cdot S_{\text{len}}^{(i)}
\end{equation}

The sampling probability $\mathbb{P}(\tau_i)$ is obtained by normalizing across the dataset:
\begin{equation}
\label{eq:ppef}
    \mathbb{P}(\tau_i) = \frac{\mathcal{Q}_i}{\sum_{\tau_j \in \mathcal{D}} \mathcal{Q}_j}
\end{equation}

Sampling via $\mathbb{P}(\tau_i)$ ensures that the sequence model is trained on high-signal, compliant, and efficient data.

\subsection{Counterfactual Regret Optimization}
\label{sec:module2}

Although CDPR enables resource perception, standard regression restricts the policy to historical averages. To achieve active improvement, we introduce the CRO mechanism. Inspired by \textit{Regret Theory}~\cite{loomes1982regret}, it utilizes a global predictor to evaluate alternative actions and identify promising local adjustments. By treating these refined outcomes as weighted regression targets, CRO directs the policy toward the empirical high-efficiency boundary, effectively overcoming mean regression.

\subsubsection{\textbf{Probabilistic Policy with Gaussian Head}}
To explore potential improvements within the high-confidence region of the data support, PRO-Bid employs a stochastic policy formulation. Drawing inspiration from~\cite{zheng2022online}, we model the bidding parameter $\lambda_t$ using a \textit{Gaussian Action Head}. Given the hidden state $h_t$ of the transformer, the policy outputs the parameters of a normal distribution:
\begin{equation}
    \pi_\theta(a_t | R_{\leq t}, C_{\leq t}, s_{\leq t}, a_{<t}) = \mathcal{N}(\mu_\theta(h_t), \sigma_\theta^2(h_t))
\end{equation}
where $\mu_\theta$ represents the nominal strategy for exploitation and $\sigma_\theta$ controls the variance for exploration.

To ground the policy within the support of the offline data and ensure stability during training, we minimize the negative log-likelihood (NLL) of the demonstrated action $a_t$:
\begin{equation}
    \mathcal{L}_{\text{NLL}} = - \log \pi_\theta(a_t | R_{\leq t}, C_{\leq t}, s_{\leq t}, a_{<t})
\end{equation}
Minimizing this loss anchors the policy mean $\mu_\theta$ to historical actions while learning appropriate variance from data noise.

\subsubsection{\textbf{Global Outcome Prediction}}
To evaluate the quality of a sampled action, the agent must anticipate its long-term impact. We introduce a \textit{Global Outcome Predictor} $\phi_\omega$, which shares the transformer backbone with the policy but utilizes a distinct head to predict cumulative future outcomes:
\begin{equation}
    (\hat{r}_{t+1:T}, \hat{c}_{t+1:T}) = \phi_\omega(R_{\leq t}, C_{\leq t}, s_{\leq t}, a_{\leq t})
\end{equation}
where $\hat{r}_{t+1:T}$ and $\hat{c}_{t+1:T}$ denote the predicted cumulative value and cost from step $t$ to the end of the episode.

Crucially, $\phi_\omega$ acts as an outcome predictor used for approximate counterfactual evaluation. By mapping state-action pairs to expected returns and costs, it enables the system to compute the constraint-aware utility for sampled alternative actions. This capability provides the foundation for the subsequent calculation of regret without necessitating online interaction.

The predictor is trained through supervised learning by minimizing the Mean Squared Error against ground-truth future values:
\begin{equation}
    \mathcal{L}_{\text{pred}} = \frac{1}{T} \sum_{t=1}^T \left( \| \hat{r}_{t+1:T} - r_{t+1:T} \|^2 + \| \hat{c}_{t+1:T} - c_{t+1:T} \|^2 \right) 
\end{equation}

\subsubsection{\textbf{Full-Episode Constraint-Aware Utility}}
A core challenge in constrained bidding is that ratio constraints apply to the entire campaign. Therefore, we define utility based on the global realized ratio by integrating historical observed data with predicted future outcomes.
Let $H_t^{r}$ and $H_t^{c}$ denote the accumulated value and cost prior to step $t$. The estimated total episode outcome for an action $a$ is $\mathcal{R}_{\text{total}}(a) = H_t^{r} + \hat{r}_{t+1:T}$ and $\mathcal{C}_{\text{total}}(a) = H_t^{c} + \hat{c}_{t+1:T}$.

The predicted global ratio is computed as $\rho(a) = \frac{\mathcal{C}_{\text{total}}(a)}{\mathcal{R}_{\text{total}}(a) + \epsilon}$.
To strongly penalize constraint violations while allowing for smooth optimization, we adopt a penalty-based scoring metric $\mathcal{P}$ to derive the final utility $U(a)$:
\begin{equation}
    \mathcal{P}(\rho(a); \rho_{\text{tgt}}) = \min \left\{ \left( \frac{\rho_{\text{tgt}}}{\rho(a)} \right)^\gamma, 1 \right\}
\end{equation}
\begin{equation}
    U(a) = \mathcal{P}(\rho(a); \rho_{\text{tgt}}) \cdot \mathcal{R}_{\text{total}}(a)
\label{eq:utility}
\end{equation}
Here, $\gamma \ge 1$ controls penalty sensitivity. This function imposes a substantial penalty on actions leading to global violations while rewarding compliant actions proportional to their total value.

\subsubsection{\textbf{Optimization via Regret-Weighted Regression}}
Inspired by \textit{Regret Theory}~\cite{loomes1982regret}, we propose a policy optimization mechanism to minimize the regret of missing higher-yield local adjustments identified via counterfactual evaluation. We adopt a robust Regret-Weighted Regression (RWR) approach. This method treats high-utility counterfactuals as regression targets, designating the deterministic policy mean $\hat{a}_t = \mu_\theta(h_t)$ as the dynamic baseline.

To estimate the potential for improvement at each time step $t$, we generate $K$ counterfactual actions $\{a_{t}^{(k)}\}_{k=1}^K$ from the current policy distribution $\pi_\theta(\cdot|h_t)$. Crucially, the counterfactual actions are sampled locally around the current policy mean $\mu_\theta$. The variance $\sigma_\theta$, learned from data noise, naturally bounds the search space to the high-confidence region of the offline data support, preventing drastic deviations into unreliable out-of-distribution (OOD) areas. We quantify the regret of omission for each action as its positive utility gain relative to the baseline:
\begin{equation}
    \Delta_{t,k} = \left[ U(a_{t}^{(k)}) - U(\hat{a}_t) \right]^+
\end{equation}
where $[\cdot]^+$ denotes the ReLU function. To ensure numerical stability and focus learning on the most promising directions, we convert these raw utility gaps into normalized importance weights using a Boltzmann distribution~\cite{rowlinson2005maxwell}:
\begin{equation}
    w_{t,k} = \frac{\exp(\Delta_{t,k} / \tau) \cdot \mathbb{I}[\Delta_{t,k} > 0]}{{\sum_{j=1}^K} \left( \exp(\Delta_{t,j} / \tau) \cdot \mathbb{I}[\Delta_{t,j} > 0] \right) + \epsilon}
\end{equation}
Here, $\tau$ is a temperature hyperparameter and $\epsilon$ prevents division by zero. This normalization bounds the gradient magnitude, effectively decoupling it from the absolute scale of utility values. The optimization goal is to shift the policy mean $\mu_\theta$ towards the weighted centroid of these high-utility actions across the entire trajectory. We formulate the loss as a weighted MSE:
\begin{equation}
    \mathcal{L}_{\text{regret}} =   \frac{1}{T} \sum_{t=1}^{T} {\sum_{k=1}^K} w_{t,k} \cdot \left\| \hat{a}_t - \text{sg}(a_{t}^{(k)}) \right\|^2 
    \label{eq:regret}
\end{equation}
where $\text{sg}(\cdot)$ indicates the stop-gradient operator. This objective provides a stable update signal, pulling the policy mean directly towards the region of higher utility identified by the counterfactuals without modifying the policy variance.

\subsubsection{\textbf{Total Training Objective}}
The final objective integrates the previously defined loss components with a Shannon entropy regularizer~\cite{haarnoja2018soft}. We explicitly formulate the total loss to highlight the specific contribution of each term:
\begin{equation}
    \mathcal{L}_{\text{total}} = \underbrace{\mathcal{L}_{\text{NLL}}}_{\text{Stability}} + \alpha \cdot \underbrace{\mathcal{L}_{\text{regret}}}_{\text{Improvement}} + \beta \cdot \underbrace{\mathcal{L}_{\text{pred}}}_{\text{Accuracy}} - \eta \cdot \underbrace{\mathcal{H}(\pi_\theta)}_{\text{Exploration}}
    \label{eq:total}
\end{equation}
where $\mathcal{H}(\pi_\theta)$ is the policy entropy. Hyperparameters $\alpha, \beta, \eta$ balance anchoring to history (Stability), optimizing utility (Improvement), ensuring predictor reliability (Accuracy), and maintaining diversity (Exploration). Crucially, this composite objective ensures that CRO functions as a constrained, local optimization based on current policy confidence, rather than an unconstrained global search, thereby mitigating the risk of drifting into unreliable OOD regions. The training procedure is summarized in Algorithm~\ref{alg:pro-bid}.

\renewcommand{\algorithmicrequire}{\textbf{Input:}}
\renewcommand{\algorithmicensure}{\textbf{Output:}}

\begin{algorithm}[t]
\caption{Training Procedure of PRO-Bid}
\label{alg:pro-bid}
\begin{algorithmic}[1]
\REQUIRE An offline training dataset $\mathcal{D}$.
\ENSURE Optimized parameters $\theta$ (Policy), $\omega$ (Predictor).

\STATE Calculate sampling distribution $\mathbb{P}(\tau)$ via Eq.~\eqref{eq:ppef}.
\FOR {Step 1,\dots, Max Step}
    \STATE Sample batch $\mathcal{B} \sim \mathbb{P}(\tau)$ and augment inputs via Eq.~\eqref{eq:aug}.
    \STATE Generate policy $\pi_\theta$ via Causal Transformer.
    \STATE Sample $K$ counterfactual actions $\{a^{(k)}\}_{k=1}^K$ from $\pi_\theta(\cdot|h_t)$.
    \STATE Predict outcomes via $\phi_\omega$ and compute utility $U$ via Eq.~\eqref{eq:utility}.
    \STATE Calculate regret loss $\mathcal{L}_{regret}$ via Eq.~\eqref{eq:regret}.
    \STATE Update $(\theta, \omega)$ to minimize the total loss $\mathcal{L}_{total}$ in Eq.~\eqref{eq:total}.
\ENDFOR
\end{algorithmic}
\end{algorithm}

%% file: chapter/4expr.tex
\newcolumntype{C}{>{\Centering\arraybackslash}X} 

\begin{table*}[ht]
\centering
\caption{Performance comparison under the metric of Score in different budget settings. The boldface denotes the highest score. The underline indicates the best result of baselines. Numerical superscript denotes the result reported in the original papers. ``*'' indicates the statistically significant improvements (i.e., two-sided t-test with p < 0.05) over the best baseline. }
\label{tab:main_table}
\renewcommand{\arraystretch}{1.2}
\resizebox{0.95\textwidth}{!}{
\begin{tabularx}{\textwidth}{ c | C | *{8}{C} c C }
\hline\hline
Dataset & Budget & BCQ & CQL & IQL & DiffBid & DT & CDT & GAS & GAVE & PRO-Bid & Improve \\
\hline
\multirow{5}{*}{AuctionNet} & 50\% & 152.8 & 171.9 & 178.2 & 77.1 & 187.1 & 175.1 & $188.4^{193}$ & $\underline{195.4}^{201}$ & \textbf{204.2*} & 4.51\% \\
 & 75\% & 210.3 & 244.3 & 246.3 & 128.4 & 272.1 & 251.1 & $278.6^{287}$ & $\underline{283.4}^{296}$ & \textbf{291.4*} & 2.82\% \\
 & 100\% & 267.4 & 289.6 & 314.1 & 189.8 & 325.8 & 328.0 & $350.9^{359}$ & $\underline{362.6}^{376}$ & \textbf{371.9*} & 2.56\% \\
 & 125\% & 313.7 & 335.7 & 351.7 & 193.5 & 394.6 & 384.5 & $404.1^{409}$ & $\underline{413.9}^{421}$ & \textbf{426.3*} & 2.99\% \\
 & 150\% & 354.5 & 363.2 & 384.6 & 234.3 & 454.9 & 437.7 & $453.4^{461}$ & $\underline{459.6}^{467}$ & \textbf{470.6*} & 2.39\% \\
\hline
\multirow{5}{*}{AuctionNet-Sparse} & 50\% & 14.77 & 16.94 & 18.16 & 12.01 & 14.88 & 17.72 & $18.52^{18.4}$ & $\underline{19.76}^{19.6}$ & \textbf{20.95*} & 6.02\% \\
 & 75\% & 21.34 & 24.27 & 24.23 & 18.63 & 22.31 & 26.39 & $27.32^{27.5}$ & $\underline{28.02}^{28.3}$ & \textbf{28.76*} & 2.64\% \\
 & 100\% & 26.32 & 28.37 & 29.16 & 23.72 & 29.96 & 32.64 & $35.72^{36.1}$ & $\underline{36.98}^{37.2}$ & \textbf{38.10*} & 3.03\% \\
 & 125\% & 32.15 & 33.20 & 36.89 & 29.78 & 34.27 & 36.36 & $39.96^{40.0}$ & $\underline{42.54}^{42.7}$ & \textbf{43.74*} & 2.82\% \\
 & 150\% & 37.52 & 37.06 & 42.19 & 37.47 & 43.73 & 44.56 & $46.57^{46.5}$ & $\underline{47.46}^{47.4}$ & \textbf{49.87*} & 5.08\% \\
\hline\hline
\end{tabularx}}
\end{table*}

\begin{table*}[ht]
\centering
\caption{Comparison across more metrics. Arrows indicate the direction of performance improvement.}
\label{tab:metric_table}
\renewcommand{\arraystretch}{1.2}
\resizebox{0.95\textwidth}{!}{
\begin{tabularx}{\textwidth}{ c | C | *{8}{C} c}
\hline\hline
Dataset & Metric & BCQ & CQL & IQL & DiffBid & DT & CDT & GAS & GAVE & PRO-Bid \\
\hline
\multirow{4}{*}{AuctionNet} & Value $\uparrow$  & 325.1 & 335.3 & 328.8 & 297.6 & 367.1 & 357.4 & 378.6 & \textbf{399.7} & 390.1 \\
 & AR $\downarrow$  & 1.020 & 0.978 & 0.871 & 1.434 & 1.003 & 0.968 & 0.966 & 1.040 & \textbf{0.862} \\
 & ER $\downarrow$  & 0.479 & 0.354 & 0.292 & 0.729 & 0.458 & 0.396 & 0.375 & 0.479 & \textbf{0.229} \\
 & Score $\uparrow$  & 267.4 & 289.6 & 314.1 & 189.8 & 325.8 & 328.0 & 350.9 & 362.6 & \textbf{371.9} \\
\hline
\multirow{4}{*}{AuctionNet-Sparse} & Value $\uparrow$  & 30.40 & 30.28 & 32.17 & 31.40 & 36.27 & 35.31 & 38.76 & \textbf{41.43} & 40.52 \\
 & AR $\downarrow$  & 0.942 & 0.922 & 0.834 & 1.358 & 0.976 & 0.857 & 0.882 & 1.014 & \textbf{0.813} \\
 & ER $\downarrow$  & 0.417 & 0.312 & 0.292 & 0.688 & 0.438 & 0.322 & 0.396 & 0.458 & \textbf{0.208} \\
 & Score $\uparrow$  & 26.32 & 28.37 & 29.16 & 23.72 & 29.96 & 32.64 & 35.72 & 36.98 & \textbf{38.10} \\
\hline\hline
\end{tabularx}}
\end{table*}

\section{Offline Experiment}
In this section, we conduct experiments on two public datasets to address the following key research questions:

\begin{itemize}[leftmargin=*]
    \item \textbf{RQ1:} \emph{ Does PRO-Bid outperform state-of-the-art baselines in terms of both value maximization and constraint adherence?}
    \item \textbf{RQ2}: \emph{How do the CDPR and CRO modules individually contribute to the overall performance of the framework? }
    \item \textbf{RQ3}: \emph{Does CDPR enable the model to perceive ratio constraint changes and maintain robustness against suboptimal data noise?}
    \item \textbf{RQ4}: \emph{Does CRO enable the policy to overcome mean regression and approach the empirical high-efficiency frontier?}
    \item \textbf{RQ5}: \emph{ How accurate is the global outcome predictor and is the policy robust against prediction errors?}
    \item \textbf{RQ6}: \emph{How sensitive is the framework to key hyperparameters?}
\end{itemize}

\subsection{Experimental Setup}
\subsubsection{\textbf{Datasets}}
We conduct experiments on AuctionNet~\cite{su2024auctionnet}, a public benchmark that provides a realistic multi-agent competitive environment for evaluation. We also employ its more challenging variant, AuctionNet-Sparse, to evaluate the model's robustness under sparse conversion signals. In all experiments, we follow the standard protocol provided by the benchmark.

\subsubsection{\textbf{Evaluation Metrics}}

We establish our evaluation framework based on the protocols of the AuctionNet benchmark. To rigorously assess bidding strategies regarding value maximization and constraint adherence, we employ four metrics:

\begin{itemize}[leftmargin=*]
    \item \textbf{Value}: The total accumulated value, such as conversion volume, obtained during the bidding period, calculated as $V = \sum_{i}x_iv_i$.
    \item \textbf{AR (Achievement Ratio)}: The ratio of the actual realized CPA to the target constraint, defined as $AR = \text{CPA}_{\text{realized}} / \allowbreak \text{CPA}_{\text{target}}$. An $AR \le 1$ indicates successful constraint satisfaction.
    \item \textbf{ER (Exceed Rate)}: A robustness metric representing the proportion of test episodes where the constraint is violated (i.e., $AR > 1$). Lower is better.
    \item \textbf{Score}: A composite metric that penalizes the total Value based on the magnitude of constraint violation. It prevents the agent from achieving high value by simply ignoring safety boundaries:
    \begin{equation}
        \text{Score} = V \times \min\left(1, \left(\frac{1}{AR}\right)^\gamma\right)
    \end{equation}
where $\gamma=2$ controls severity of the penalty for overspending.
\end{itemize}

\subsubsection{\textbf{Baselines}}
We compare our method to a wide range of baselines involving offline RL, diffusion models, and decision transformer approaches. For offline RL methods, BCQ~\cite{fujimoto2019off}, which constrains Q-learning to remain close to the behavior policy; CQL~\cite{kumar2020conservative}, which learns a conservative Q-function to penalize out-of-distribution actions; IQL~\cite{kostrikov2021offline}, which avoids explicit behavior cloning by learning implicit in-sample value functions. For diffusion-based methods, DiffBid~\cite{guo2024generative}, a conditional diffusion model that generates bidding trajectories from contextual features. For transformer-based methods, DT~\cite{chen2021decision}, which  employs a transformer architecture for sequential decision-making modeling; CDT~\cite{liu2023constrained}, a DT-based approach that handles a vector of multiple constraints; GAS~\cite{li2025gas}, a DT-based framework with post-training search by applying Monte Carlo Tree Search (MCTS); GAVE~\cite{gao2025generative}, which guides DT through Score-based RTG and a learnable value function.

\subsubsection{\textbf{Implementation Details}}
All experiments are conducted on a compute cluster equipped with NVIDIA L20 GPUs. The model is trained for a total of 200,000 steps using a fixed batch size of 128. 
Consistent with the architecture of state-of-the-art baselines, the backbone of PRO-Bid is instantiated as a Causal Transformer featuring 8 attention layers and 16 attention heads. 
We optimize the model parameters using the AdamW optimizer with a learning rate of $1 \times 10^{-5}$. 
To ensure a rigorous and fair comparison, we strictly align the experimental settings across all methods. Specifically, all transformer-based baselines share the identical backbone capacity to eliminate parameter scale bias, and all methods utilize the exact same offline datasets and evaluation protocols. Furthermore, we adopt the default hyperparameters suggested in their respective papers and carefully fine-tune them to ensure they are evaluated at their optimal performance.
To guarantee statistical significance, we conduct 10 independent runs with different random seeds and report the average performance metrics.

\subsection{Overall Performance (RQ1)}
We analyze the results in Table~\ref{tab:main_table} and Table~\ref{tab:metric_table}. The comparative evidence supports three critical conclusions regarding value maximization and constraint satisfaction.

\textbf{PRO-Bid consistently achieves superior performance across diverse budgetary scenarios.} As shown in Table~\ref{tab:main_table}, PRO-Bid obtains the highest Score across all budget settings. To ensure a rigorous evaluation, we denote results reported in original papers with superscripts where our reproduction yielded lower figures.
Even compared to these stronger original baselines, PRO-Bid maintains a consistent performance advantage. This lead is particularly evident on the challenging AuctionNet-Sparse dataset. These results confirm the robustness of our policy against data sparsity and varying constraints without overfitting to specific historical distributions.

\textbf{PRO-Bid demonstrates high adherence to hard constraints.} Table~\ref{tab:metric_table} reveals a critical trade-off between  raw conversion volume and constraint satisfaction. While certain baselines achieve marginally higher conversion totals, they often violate the target CPA, indicated by an AR that exceeds the safety threshold. In contrast, PRO-Bid maintains the AR strictly within the feasible region while achieving the lowest ER. This confirms that our framework operates precisely within safety boundaries, ensuring that value maximization does not occur at the expense of financial risk.

\textbf{PRO-Bid establishes an optimal balance between safety and efficiency.} 
Our approach outperforms both conservative reinforcement learning baselines and aggressive generative methods. Standard reinforcement learning algorithms tend to exhibit excessive conservatism to avoid out-of-distribution actions, yielding suboptimal value. Conversely, naive generative approaches often lack precise resource perception, leading to aggressive bidding that breaches constraints. PRO-Bid effectively addresses this dichotomy by dynamically adjusting bidding aggressiveness based on resource consumption, thereby converging towards the empirical high-efficiency frontier of compliance and efficiency.

\subsection{Ablation Study (RQ2)}
To quantify the contribution of each component, we systematically remove or degrade the CDPR and CRO modules. To explicitly evaluate the internal mechanisms of CDPR, we separate the dual-stream context and the Pareto reweighting. The fine-grained results in Table~\ref{tab:ablation} demonstrate the distinct role of each design choice.

\textbf{CDPR provides the foundation for constraint perception and data quality.} The complete removal of CDPR (\textit{w/o CDPR}) results in the most severe degradation in constraint satisfaction across all metrics. This confirms that without the foundational perception and quality control provided by CDPR, the model cannot effectively navigate the complex trade-offs in constrained bidding. 
Specifically, removing the cost-to-go conditioning (\textit{w/o Dual}) leads to a significant increase in the Exceed Rate and Achievement Ratio. This indicates that without explicitly decoupling cost and value, the agent suffers from state aliasing and fails to track resource boundaries, resulting in frequent overspending. Meanwhile, replacing the Pareto-prioritized reweighting with uniform sampling (\textit{w/o Pareto}) causes a decline in conversion volume and total Score. This suggests that uniform sampling over noisy historical logs forces the model to mimic suboptimal strategies, whereas our reweighting mechanism successfully distills high-quality demonstrations from the dataset.

\textbf{CRO elevates policy performance by targeting high-utility regions.} The exclusion of CRO results in a notable decline in both conversion volume and score, whereas constraint adherence remains relatively stable. This indicates that while the base model adheres to safety boundaries, it encounters difficulties in maximizing returns within those limits. CRO functions as an active optimization engine. By regressing the policy mean toward the weighted centroid of high-utility local adjustments, it effectively smooths out individual predictor errors and pulls the strategy toward the empirical high-efficiency frontier.

\begin{table}[hbtp]
\centering
\caption{Ablation Study with 100\% budget. \textit{w/o Dual} removes the cost-to-go conditioning in CDPR; \textit{w/o Pareto} replaces Pareto-prioritized reweighting with uniform sampling.}
\label{tab:ablation}
\setlength{\tabcolsep}{3.5pt}
\renewcommand{\arraystretch}{1.1}
\begin{tabular}{@{}lcccccccc@{}}
\toprule
\multirow{2}{*}{Method} &
\multicolumn{4}{c}{AuctionNet} &
\multicolumn{4}{c}{AuctionNet-Sparse} \\
\cmidrule(r){2-5} \cmidrule(l){6-9}
& Conv & AR & ER & Score & Conv & AR & ER & Score \\
\midrule
PRO-Bid  & \textbf{390} & \textbf{0.86} & \textbf{0.23} & \textbf{372} & \textbf{40.5} & \textbf{0.81} & \textbf{0.21} & \textbf{38.1} \\
\hline
w/o Dual   & 375   & 0.91   & 0.29   & 356   & 39.0   & 0.84   & 0.29   & 35.5  \\
w/o Pareto & 378   & 0.86   & 0.23   & 363   & 39.2   & 0.83   & 0.21   & 36.9   \\
\hline
w/o CDPR & 372 & 0.94 & 0.35 & 347 & 37.3 & 0.92 & 0.31 & 34.6 \\
w/o CRO & 380 & 0.89 & 0.27 & 361 & 39.4 & 0.85 & 0.25 & 36.7 \\
\bottomrule
\end{tabular}
\end{table}

\subsection{Effectiveness of CDPR (RQ3)}
\begin{figure}[hbtp]
    \centering
    \includegraphics[width=0.98\linewidth]{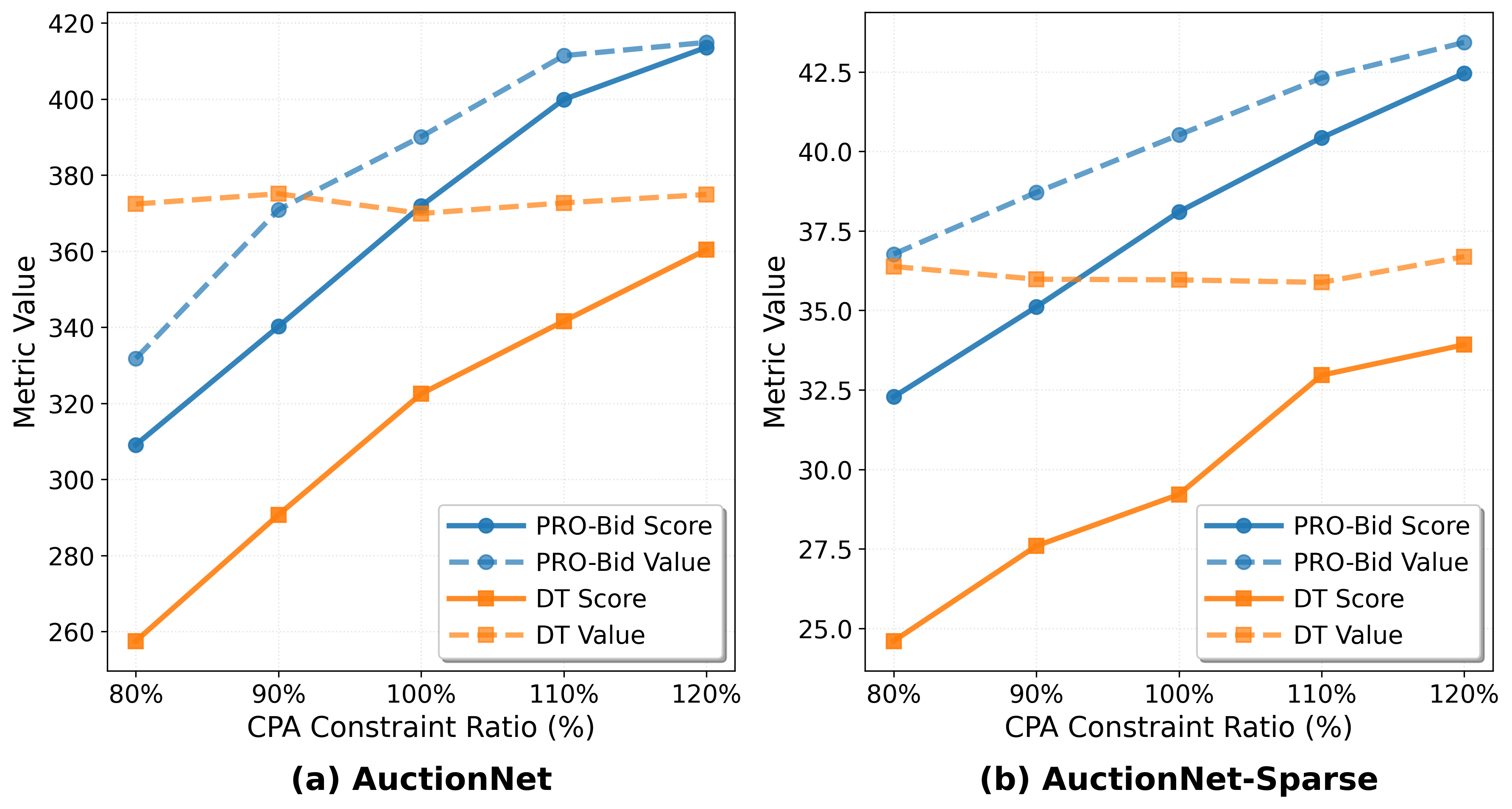}
    \caption{Comparison in different CPA constraint settings.}
    \label{fig:cpa}
\end{figure}

\begin{figure}[hbtp]
    \centering
    \includegraphics[width = 0.98\linewidth]{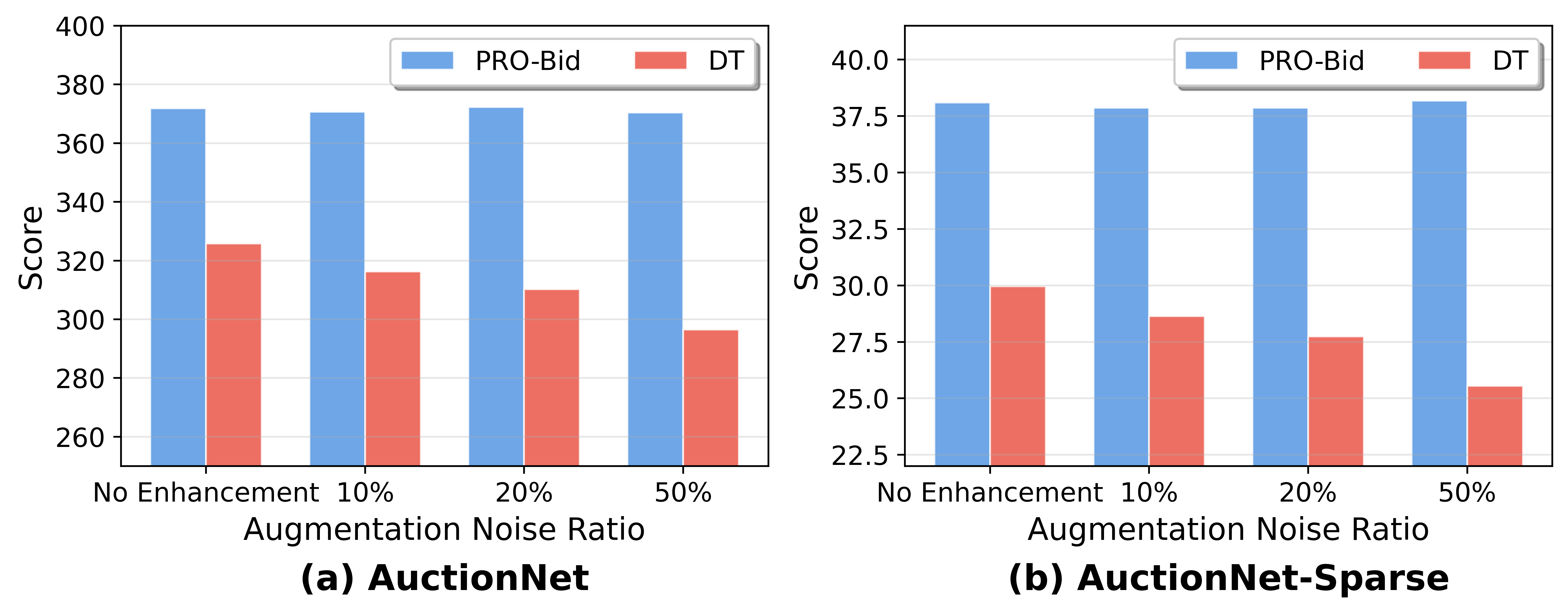}
    \caption{Performance under different noise augmentation.}
    \label{fig:noise}
\end{figure}

To answer RQ3, we investigate whether PRO-Bid successfully perceives variations in constraints and resists data noise. We conduct a dual-perspective analysis using quantitative sensitivity tests presented in Figure~\ref{fig:cpa} and robustness tests shown in Figure~\ref{fig:noise}.

\textbf{CDPR enables the agent to adaptively adjust bidding aggressiveness based on varying constraint targets.} Figure~\ref{fig:cpa} contrasts model behaviors under different CPA constraints. PRO-Bid demonstrates a clear positive correlation between constraint relaxation and performance metrics. This indicates that the decoupled context effectively resolves state aliasing, which allows the policy to distinguish between tight and loose feasible regions. Consequently, the agent adaptively scales its bidding intensity to maximize value within a specific allowance. In contrast, the baseline displays a rigid performance curve, confirming that static ratio conditioning fails to capture shifting resource boundaries. This confines the agent to suboptimal average behaviors regardless of the actual target.

\textbf{CDPR ensures robustness against suboptimal data quality through Pareto-based reweighting.} We empirically validate the resilience of the model by training on datasets with increasing ratios of synthetic noise, as illustrated in Figure~\ref{fig:noise}. The results expose a critical vulnerability in standard methods, where uniform sampling leads to significant performance degradation as noise levels rise. PRO-Bid exhibits remarkable robustness, maintaining consistent high scores across all noise settings. This confirms that the Pareto-prioritized reweighting effectively neutralizes data contamination, allowing the model to distill efficient policies even from highly polluted industrial logs where standard approaches collapse.

\subsection{Effectiveness of CRO (RQ4)}
\begin{figure}[hbtp]
    \centering
    \includegraphics[width=0.98\linewidth]{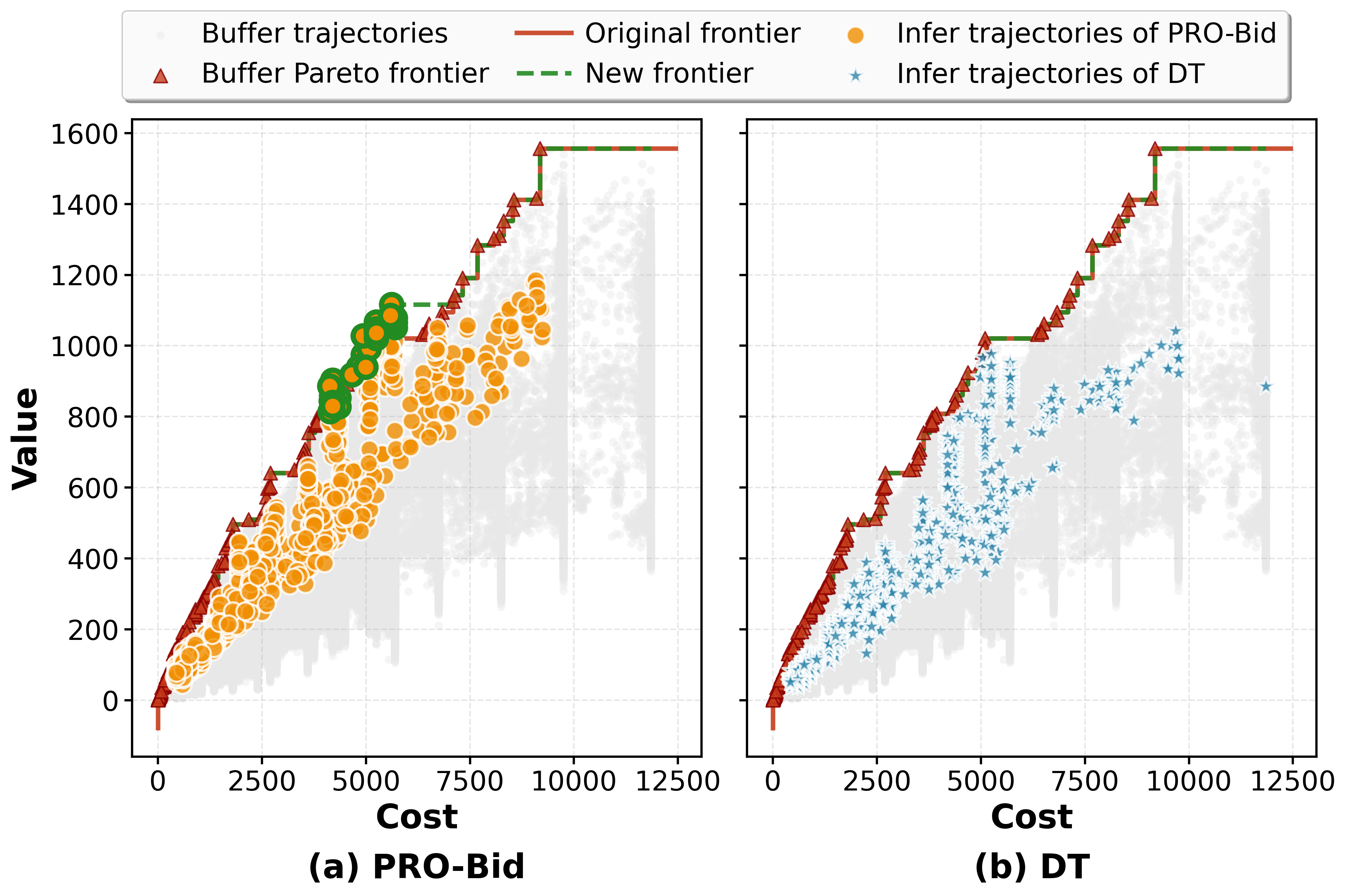}
    \caption{Visualization of inference results on AuctionNet.}
    \label{fig:beyond}
\end{figure}

To answer RQ4, we visualize the inference trajectories in the objective space relative to the empirical Pareto frontier (Figure~\ref{fig:beyond}).

\textbf{CRO shifts the policy toward the high-efficiency boundary via bounded local refinement.} The visualization highlights a clear behavioral divergence. Standard DT remains trapped within the dense interior of the historical data, mimicking average behaviors. Conversely, PRO-Bid demonstrates a stark distributional shift toward the empirical frontier.
While some inferred trajectories extend beyond the historical frontier, this stems from local interpolation rather than unconstrained extrapolation. The empirical frontier merely forms a discrete envelope of finite logs. Trained on the entire dataset, the predictor $\phi_\omega$ learns a smoothed return-cost landscape. By evaluating actions strictly within the policy's learned variance, CRO performs a safe local search. This mechanism allows the agent to identify and compose better local trade-offs within the continuous manifold of the data support, effectively improving upon discrete historical bounds without leaving the high-confidence region.

\subsection{Predictor Reliability and Robustness (RQ5)}
\label{sec:rq5}

To answer RQ5, we investigate the reliability of the global outcome predictor and the resilience of the policy under prediction noise. We conduct a dual-perspective analysis using the prediction errors shown in Table~\ref{tab:predictor_accuracy} and the robustness tests presented in Table~\ref{tab:predictor_noise}.

\textbf{The global outcome predictor achieves reliable estimation accuracy.} As shown in Table~\ref{tab:predictor_accuracy}, the predictor maintains consistently low error rates in terms of both RMSE and MAE for value and cost estimation across different benchmark environments. This confirms that the transformer backbone effectively captures the underlying environment dynamics despite the inherent variance in advertising logs. Consequently, the module provides a solid foundation for evaluating the utility of sampled alternative actions without requiring actual online interactions.

\begin{table}[t]
\centering
\caption{Prediction accuracy of the global outcome predictor.}
\label{tab:predictor_accuracy}
\setlength{\tabcolsep}{6pt}
\renewcommand{\arraystretch}{1.1}
\begin{tabular}{@{}lcccc@{}}
\toprule
\multirow{2}{*}{Metric} &
\multicolumn{2}{c}{AuctionNet} &
\multicolumn{2}{c}{AuctionNet-Sparse} \\
\cmidrule(r){2-3} \cmidrule(l){4-5}
& Value & Cost & Value & Cost \\
\midrule
RMSE & 7.59 & 52.87 & 0.79 & 23.02 \\
MAE  & 5.42 & 36.04 & 0.41 & 12.06 \\
\bottomrule
\end{tabular}
\end{table}

\textbf{PRO-Bid exhibits graceful degradation against prediction errors.} Table~\ref{tab:predictor_noise} contrasts model behaviors under varying levels of multiplicative noise applied to the predicted outcomes. PRO-Bid demonstrates stable constraint adherence, showing only marginal Score drops and slight increases in violations as perturbation grows. The robustness is primarily attributed to the regret-weighted regression mechanism in CRO. By shifting the policy mean toward the weighted centroid of multiple high-utility actions, the framework effectively smooths out individual prediction errors, preventing the agent from overreacting to isolated faulty estimates.

\begin{table}[t]
\centering
\caption{PRO-Bid performance under predictor perturbation with $\pm 5\%$ and $\pm 10\%$ multiplicative noise.}
\label{tab:predictor_noise}
\setlength{\tabcolsep}{4pt}
\renewcommand{\arraystretch}{1.1}
\begin{tabular}{@{}lcccccc@{}}
\toprule
\multirow{2}{*}{Perturbation} &
\multicolumn{3}{c}{AuctionNet} &
\multicolumn{3}{c}{AuctionNet-Sparse} \\
\cmidrule(r){2-4} \cmidrule(l){5-7}
& Score & AR & ER & Score & AR & ER \\
\midrule
origin     & 371.9 & 0.862 & 0.229 & 38.10 & 0.813 & 0.208 \\
$\pm 5\%$  & 368.7   & 0.881  & 0.250  & 37.82 & 0.857 & 0.229 \\
$\pm 10\%$ & 364.4  & 0.893  & 0.271  & 37.01 & 0.876 & 0.271 \\
\bottomrule
\end{tabular}
\end{table}

\subsection{Hyperparameter Sensitivity (RQ6)}

\begin{figure}[htbp]
    \centering
    \includegraphics[width=\linewidth]{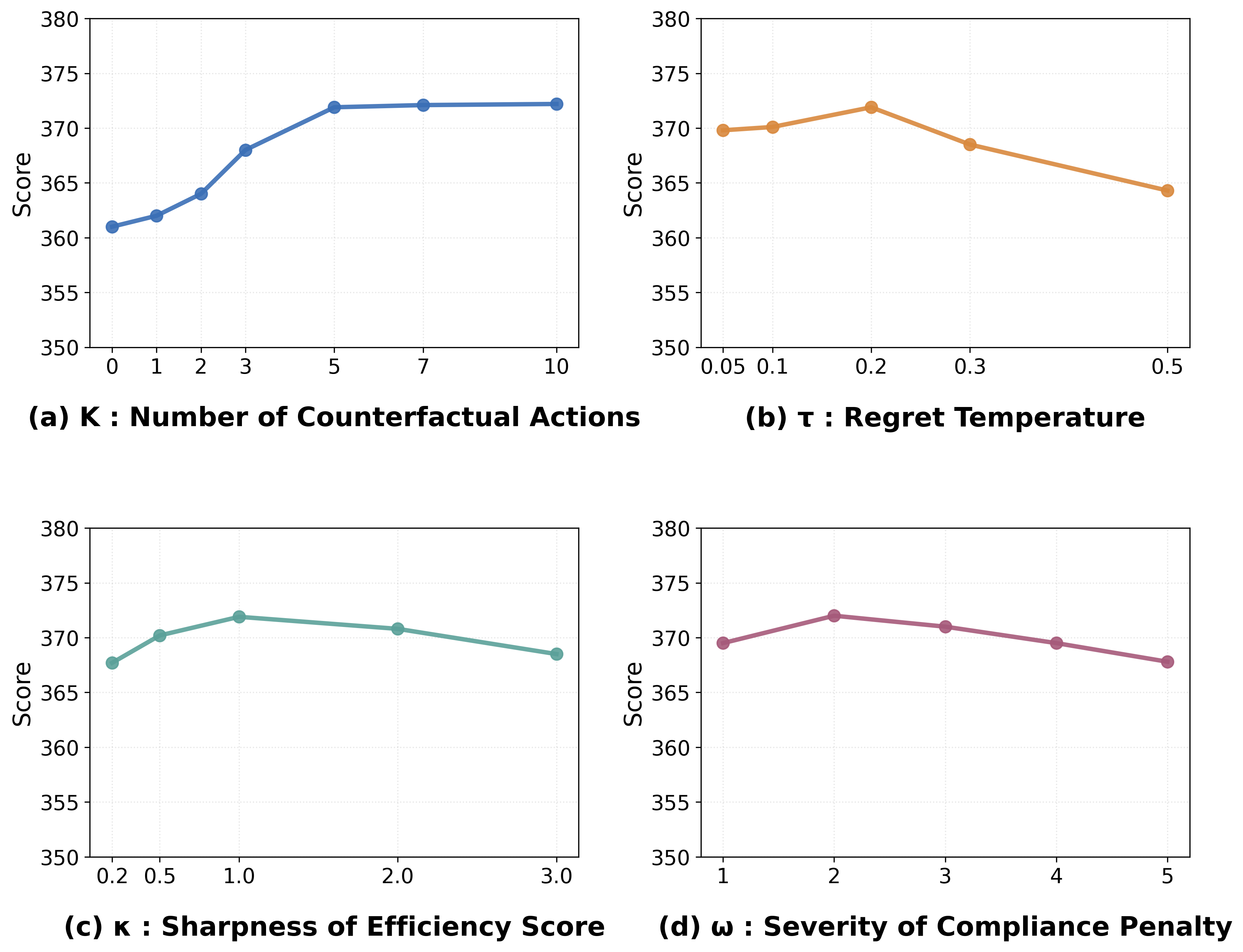}
    \caption{Sensitivity analysis of key hyperparameters on AuctionNet.}
    \label{fig:hyperparam}
\end{figure}

To answer RQ6, we investigate the sensitivity of PRO-Bid to key hyperparameters across both the CRO and CDPR modules. Figure~\ref{fig:hyperparam} illustrates the performance variations on the AuctionNet.

\textbf{CRO parameters balance the search space and optimization focus.} Figures~\ref{fig:hyperparam}(a) and (b) evaluate the counterfactual sample size $K$ and regret temperature $\tau$. Increasing $K$ expands the search area, improving performance until the high-yield region is adequately covered and marginal gains diminish. For $\tau$, an intermediate value is strictly preferred. Setting $\tau$ too low concentrates all regression weights on a single action and harms training stability. Conversely, an excessively large $\tau$ flattens the distribution and fails to prioritize high-utility adjustments.

\textbf{CDPR parameters balance data quality and trajectory diversity.} Figure~\ref{fig:hyperparam} (c) and (d) evaluate the shape parameters $\kappa$ and $\omega$ in the Pareto reweighting. Both exhibit an inverted-U trend, achieving optimal performance at moderate values. This highlights a critical trade-off: overly strict efficiency reweighting or severe compliance penalties assign near-zero weights to too many trajectories, which damages the representation capacity of the dataset. Conversely, insufficient penalties allow suboptimal logs to contaminate the learning process.

%% file: chapter/5online.tex
\section{Online A/B Test}
\begin{figure}[t]
    \centering
    \includegraphics[width = 0.98\linewidth]{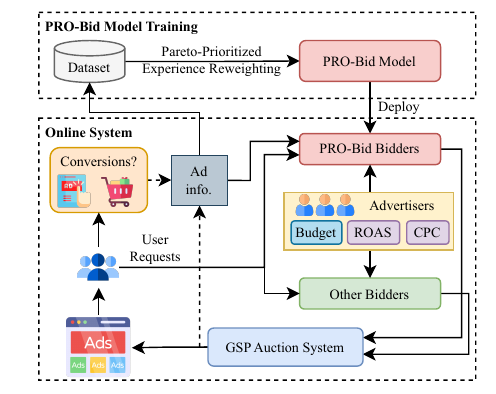}
    \caption{Online Auto-bidding System.}
    \label{fig:online}
\end{figure}

To evaluate the practical effectiveness of PRO-Bid, we deployed it on the advertising system of a large-scale e-commerce platform, within the bidding framework as illustrated in Figure~\ref{fig:online}. In this scenario, advertisers specify total budgets along with optional efficiency constraints, such as CPC or ROAS, requiring the bidding strategy to maximize traffic value while maintaining robust compliance with these financial limits. Following the generalizability of the CDPR module, the online policy seamlessly adapts to these requirements by redefining the RTG. We compared PRO-Bid against the production baseline, which utilizes a highly optimized RL approach. The deployment covers a wide range of advertising categories to validate the framework's versatility across different market segments. The details of deployment are provided in Appendix~\ref{app:deploy}, and the experimental setup is outlined as follows:

\begin{itemize}[leftmargin=*]
    \item \textbf{Action}: The model outputs the bidding parameter $\lambda_t$ at each time step $t$, which is used to dynamically adjust bid prices.
    \item \textbf{State}: The feature vector encapsulates campaign dynamics including remaining time, budget consumption speed, deviation from targets, and historical market statistics.
    \item \textbf{Return-to-Go}: Given the sparsity and delay associated with real-time signals, the RTG is defined as the cumulative sum of expected GMV from the current time step to the end of the day.
    \item \textbf{Cost-to-Go}: The Cost-to-Go represents the remaining budget, allowing the model to explicitly track financial boundaries.
\end{itemize}

Online A/B testing was conducted over a consecutive seven-day period. To ensure a rigorous evaluation, we utilized a randomized bucket testing mechanism, allocating 10\% of the total platform traffic to both the control and experimental groups. Due to commercial confidentiality constraints, we report only standard business metrics and their relative improvements compared to the baseline. The results are summarized in Table~\ref{tab:online_results}, where $\mathrm{CONS_{rate}}$ denotes the constraint compliance rate. PRO-Bid consistently exhibits robust adaptability, achieving statistically significant improvements over the baseline, including a 7.07\% increase in GMV, a 9.56\% increase in clicks, and a 7.98\% increase in ROAS. Notably, these performance gains are accompanied by a 6.18\% enhancement in constraint satisfaction, which validates the capability of PRO-Bid to balance aggressive value acquisition with strict adherence to constraints. These results confirm the robust practical viability of PRO-Bid for large-scale production systems characterized by complex market dynamics and diverse marketing requirements.

\begin{table}[h]
\centering
\caption{Online A/B test results.}
\label{tab:online_results}
\begin{tabular}{lcccc}
\toprule
Metric & GMV & CLICK & ROAS & $\mathrm{CONS_{rate}}$ \\
\midrule
Improve & +7.07\% & +9.56\% & +7.98\% & +6.18\% \\
\bottomrule
\end{tabular}
\end{table}

%% file: chapter/2related.tex
\section{Related Work}

\subsection{Offline Reinforcement Learning}
Reinforcement learning has evolved from online algorithms like policy gradients \cite{sutton1999policy} and deep Q-Learning \cite{mnih2013playing} to offline paradigms that eliminate the safety risks and costs of real-world interaction \cite{kiyohara2021accelerating}. A central challenge in this domain is the mitigation of the distributional shift between the learned policy and the behavioral policy. Value-based methods offer robust solutions to this problem: BCQ \cite{fujimoto2019off} restricts actions to the dataset support, CQL \cite{kumar2020conservative} penalizes out-of-distribution values, and IQL \cite{kostrikov2021offline} employs expectile regression to estimate values without querying unseen actions. IDQL \cite{hansen2023idql} further improves robustness by integrating diffusion models with implicit value learning. However, these methods based on the Markov Decision Process (MDP) often encounter difficulties in capturing the long-range dependencies inherent in sequential tasks.  The Decision Transformer \cite{chen2021decision} addresses this limitation by recasting RL as a sequence modeling problem, generating actions autoregressively from historical trajectories to eliminate bootstrapping errors. Extensions like CDT \cite{liu2023constrained} further integrate safety tokens, balancing performance with strict boundaries.

\subsection{Auto-Bidding}
Auto-bidding serves as the core infrastructure for programmatic advertising, optimizing key performance indicators under strict budget constraints \cite{deng2021towards,aggarwal2024auto}. While early control-theoretic approaches like PID \cite{chen2011real} and OnlineLP \cite{yu2017online} provided efficient pacing, they rely on simplified assumptions ill-suited for dynamic markets. Consequently, deep RL methods like RLB \cite{cai2017real}, USCB \cite{he2021unified}, MAAB \cite{wen2022cooperative} and SORL \cite{mou2022sustainable} were developed to handle high-dimensional states, with the industry later gravitating towards offline frameworks such as BCQ \cite{fujimoto2019off} and IQL \cite{kostrikov2021offline} to mitigate the financial risks of online exploration. However, these MDP-based methods often struggle with long-horizon budget allocation, prompting a shift towards generative sequence modeling. Conditional diffusion models like DiffBid \cite{guo2024generative} and EGDB \cite{peng2025expert} generate holistic trajectories, with the latter incorporating expert guidance to enhance planning. Similarly, transformer-based architectures have evolved to address specific offline challenges: GAS \cite{li2025gas} utilizes post-training search, GAVE \cite{gao2025generative} introduces value-guided exploration to prevent behavioral collapse, and EBaReT \cite{li2025ebaret} tackles data quality and reward sparsity through expert-guided inference. These generative paradigms collectively demonstrate superior capability in capturing complex temporal correlations compared to traditional step-wise decision-making.

%% file: chapter/6conclu.tex
\section{Conclusion}
We propose PRO-Bid, a generative framework designed for constrained auto-bidding through systematic redesigns across data, architecture, and training. By integrating CDPR and CRO, our approach effectively addresses the challenges of state aliasing and the limitations of regression-based learning. Rather than mimicking average historical behaviors, PRO-Bid utilizes regret-weighted targets to actively guide the policy toward the empirical high-efficiency frontier. Extensive offline and online experiments demonstrate that PRO-Bid achieves superior value acquisition while maintaining high compliance with efficiency constraints, establishing a robust solution for auto-bidding in complex advertising environments.

\section{Limitation and Future Work}
\label{sec:limitation}
While PRO-Bid restricts counterfactual sampling within the policy's learned variance, its deterministic utility evaluation remains a limitation. The global outcome predictor $\phi_\omega$, trained strictly on historical logs, may introduce subtle covariate shifts when evaluating actions near the empirical Pareto frontier in dynamic markets. Without explicit epistemic uncertainty quantification, point estimates can occasionally overestimate the utility of boundary actions. Future work could integrate uncertainty-aware modeling or pessimistic value estimation to penalize unreliable predictions without compromising online inference latency.

%% file: chapter/7appendix.tex
\section{Deploy Details of Online A/B Test}
\label{app:deploy}

To adapt to the evolving e-commerce environment, we employ a rolling window training strategy using logs from the most recent 14 consecutive days, comprising approximately 300,000 bidding trajectories. To handle the long-tail distribution of real-world data, we apply logarithmic normalization to different feature categories. The utility function in Regret-Weighted Regression jointly weighs value acquisition and constraint satisfaction, punishing violations while rewarding high-yield outcomes. 

During online inference, the initial Cost-to-Go is set to the total daily budget, and the initial Return-to-Go is derived from the specific efficiency constraint; both tokens are dynamically updated from realized feedback as real-time pacing signals. To satisfy strict low-latency requirements, we use the deterministic mean of the output action distribution as the final bidding parameter $\lambda_t$, avoiding the computational overhead of sampling.